\documentclass[conference]{IEEEtran}
\IEEEoverridecommandlockouts
\usepackage{cite}
\usepackage[colorlinks,
linkcolor=black,
anchorcolor=black,
citecolor=black,
urlcolor=black
]{hyperref}
\usepackage[english]{babel}
\usepackage{graphicx}
\graphicspath{{Figures/}}
\DeclareGraphicsExtensions{.pdf,.jpeg,.png}
\usepackage{amsmath,amssymb}
\usepackage{gensymb}
\usepackage{xcolor,colortbl,arydshln}
\usepackage{lipsum}
\usepackage[mathlines,switch]{lineno}
\usepackage{algpseudocode, algorithm}
\usepackage{multirow}
\usepackage{colortbl}
\usepackage{xcolor}
\usepackage{mathrsfs}
\usepackage{nomencl}
\usepackage[cal=boondox]{mathalfa}
\usepackage{ragged2e}
\usepackage{makecell, boldline}
\definecolor{cerulean}{rgb}{0.0, 0.48, 0.65}
\definecolor{grn}{rgb}{0, .8, .6}
\definecolor{brickred}{rgb}{0.8, 0.25, 0.33}
\def\BibTeX{{\rm B\kern-.05em{\sc i\kern-.025em b}\kern-.08em
    T\kern-.1667em\lower.7ex\hbox{E}\kern-.125emX}}
\begin{document}

\title{Reinforcement Learning based Dynamic Model Selection for Short-Term Load Forecasting}

\author{\IEEEauthorblockN{Cong Feng, Jie Zhang}
\IEEEauthorblockA{
\textit{The University of Texas at Dallas}\\
\{cong.feng1, jiezhang\}@utdallas.edu}
}

\maketitle

\begin{abstract}
With the growing prevalence of smart grid technology, short-term load forecasting (STLF) becomes particularly important in power system operations. There is a large collection of methods developed for STLF, but selecting a suitable method under varying conditions is still challenging. This paper develops a novel reinforcement learning based dynamic model selection (DMS) method for STLF. A forecasting model pool is first built, including ten state-of-the-art machine learning based forecasting models. Then a Q-learning agent learns the optimal policy of selecting the best forecasting model for the next time step, based on the model performance. The optimal DMS policy is applied to select the best model at each time step with a moving window. Numerical simulations on two-year load and weather data show that the Q-learning algorithm converges fast, resulting in effective and efficient DMS. The developed STLF model with Q-learning based DMS improves the forecasting accuracy by approximately 50\%, compared to the state-of-the-art machine learning based STLF models.
\end{abstract}

\begin{IEEEkeywords}
Q-learning, reinforcement learning, model selection, load forecasting, machine learning
\end{IEEEkeywords}

\section{Introduction}
Accurate short-term load forecasting (STLF) plays an increasingly important role in reliable and economical power system operations. For example, STLF can be used for calculating load baselines in the design of demand response program~\cite{chen2017short}. STLF can also be used in the real-time unit commitment and economic dispatch~\cite{saksornchai2005improve}, extreme ramping event detection, energy trading, energy storage management, etc~\cite{cui2017characterizing}. 

To improve the load forecasting accuracy, a large number of methods have been developed in the past decades. STLF methods can be classified into different categories based on forecasting time horizons, spatial scales, and method principles. According to the forecasting method principle, STLF methods can be roughly categorized into statistical methods, machine learning methods, and deep learning methods. Statistical methods are usually built based on only the historical time series. The most popular used STLF methods are machine learning based models, which are able to integrate external information such as meteorological data. Deep learning methods have also been recently applied to STLF. For example, a pooling deep recurrent neural network was developed, which has shown to outperform Autoregressive integrated moving average model and support vector regression (SVR) model by 19.5\% and 13.1\%, respectively~\cite{shi2017deep}. A more comprehensive review of STLF methods can be found in recent review papers~\cite{raza2015review, yildiz2017review}.

To better utilize the various STLF methods, research has been done to further improve the forecasting accuracy by performing model selection under varying conditions. First, ensemble forecasting methods have been developed to select and combine multiple models. For instance, Alamaniotis~\textit{et al.}~\cite{alamaniotis2012evolutionary} linearly combined six Gaussian processes (GPs), which outperformed individual GPs. Feng~\textit{et al.}~\cite{feng2018short} aggregated multiple artificial neural network (ANN), SVR, random forest (RF), and gradient boosting machine (GBM) models to mitigate the risk of choosing unsatisfactory models. Second, load patterns are classified into clusters based on some similarities to select the best load forecasting model in each cluster. For example, Wang~\textit{et al.}~\cite{wang2016factors} developed a \textit{K}-means-based least squares SVR (LS-SVR) STLF method, which clustered load profiles using  \textit{K}-means and forecasted load using an LS-SVR model in each cluster. Though these strategies help improve the forecasting accuracy, it is still challenging to select the best forecasting model at each forecasting time step.

In this paper, a Q-learning based dynamic model selection (DMS) framework is developed, which aims to choose the best forecasting model from a pool of state-of-the-art machine learning models at each time step. The main contributions of this paper include: (i) building an STLF model pool based on state-of-the-art machine learning algorithms; (ii) developing a Q-learning based DMS framework to determine the best forecasting model at each forecasting time step; (iii) improving the forecasting accuracy by approximately 50\%.

The remainder of this paper is organized as follows. Section~\ref{section2} presents the developed STLF method with Q-learning DMS (M$_Q$). Case studies and result analysis are discussed in Section~\ref{section3}. Concluding remarks and future work are given in Section~\ref{section4}.

\section{STLF with Reinforcement Learning based Dynamic Model Selection}\label{section2}
\begin{figure*}[!ht]
	\centering
	\includegraphics[width =4.88in]{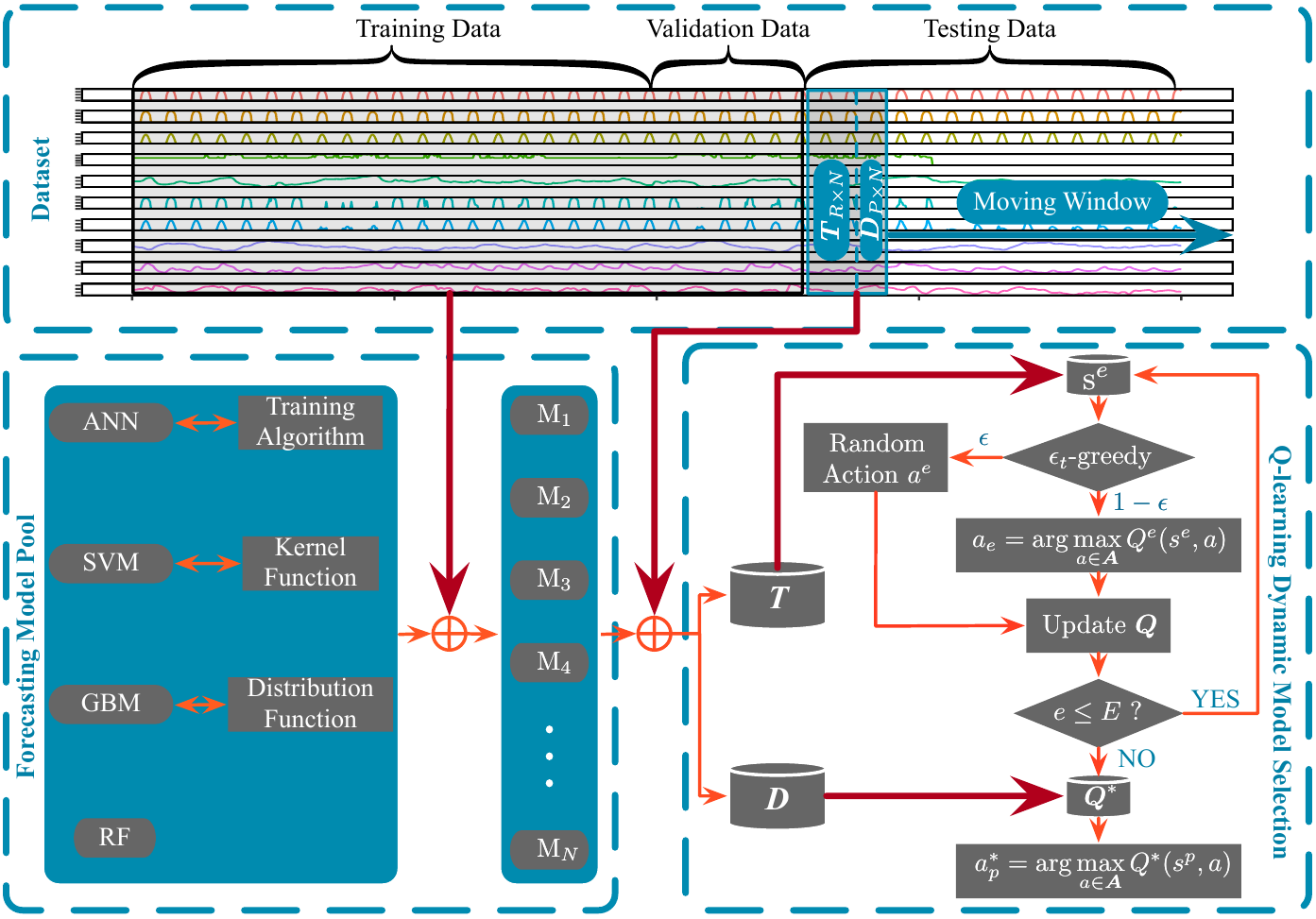}
	\caption{The framework of the developed STLF method with Q-learning based Dynamic Model Selection (DMS) (M$_Q$).}
	\label{framework}\vspace{- .18 in}
\end{figure*}

In this section, the developed STLF method with reinforcement learning based DMS is described. The overall framework of the STLF with Q-learning based DMS (M$_Q$) is illustrated in Fig.~\ref{framework}, which consists of two major components: (i) a forecasting model
pool with ten candidate forecasting models, and (ii) Q-learning based DMS.

\subsection{STLF Machine Learning Model Pool}
A collection of machine learning based STLF models constitute the forecasting model pool, from which the best model is selected at each time step in the forecasting stage. The model pool consists of ten models with four machine learning algorithms diversified by different training algorithms, kernel functions, or distributions functions as shown in the bottom left box of Fig.~\ref{framework}. Specifically, three ANN models with standard back-propagation (BP), momentum-enhanced BP, and resilient BP training algorithms are selected based on their fast convergence and satisfactory performance.	The most popular kernels in SVR are used, which are linear, polynomial, and radial base function kernels. GBM models with squared, Laplace, and T-distribution loss functions are empirically selected. The last model is an RF model. The details of the models are summarized in Table~\ref{kernel}. These models are trained based on a training dataset, and hyperparameters are tuned using a validation dataset (a list of hyperparameters could be found in Ref.~\cite{feng2017data}). 
\begin{table}[!htb]
	\caption{Machine Learning based Forecasting Model Pool}
	\label{kernel}
	\begin{center}  
		\begin{tabular}{clcccc}  
			\rowcolor{cerulean} \textcolor{white}{\textbf{Algorithm}}&\textcolor{white}{\textbf{Model}}&\textcolor{white}{\textbf{Training algorithm or function}}\\
			\multirow{3}{*}{\textbf{ANN}}
			&\multicolumn{1}{l|}{M$_1$}&Standard back-propagation\\
			&\multicolumn{1}{l|}{M$_2$}&Momentum-enhanced back-propagation\\
			&\multicolumn{1}{l|}{M$_3$}&Resilient back-propagation\\
			\hline	\multirow{3}{*}{\textbf{SVM}}
			&\multicolumn{1}{l|}{M$_4$}&Radial basis function kernel\\
			&\multicolumn{1}{l|}{M$_5$}&Linear kernel\\
			&\multicolumn{1}{l|}{M$_6$}&Polynomial kernel\\
			\hline	\multirow{3}{*}{\textbf{GBM}}
			&\multicolumn{1}{l|}{M$_7$}&Squared loss\\
			&\multicolumn{1}{l|}{M$_8$}&Laplace loss\\
			&\multicolumn{1}{l|}{M$_9$}&T-distribution loss\\
			\hline	\multirow{1}{*}{\textbf{RF}}
			&\multicolumn{1}{l|}{M$_{10}$}&CART aggregation\\
			\hline 
		\end{tabular}\vspace{-.22 in}
	\end{center} 
\end{table}
\subsection{Q-learning based Dynamic Model Selection (DMS)}
Once forecasts are independently generated by forecasting models in the model pool, the best model is selected by a reinforcement learning agent at each forecasting time step. Reinforcement learning is a typical machine learning algorithm that models an agent interacting with its environment. In this paper, Q-learning, a model-free adaptive dynamic programming algorithm, is adopted to learn the optimal policy of finding the best forecasting model at every forecasting time step. 

In order to train the Q-learning agent, a mathematical framework of DMS is first defined in a Markov Decision Process (MDP). In general, a Q-learning agent takes sequential \textit{actions} at a series \textit{states} based on a state-action value matrix \textit{Q-table} until reaching an ultimate goal~\cite{yan2017q}. The actions are evaluated by a scalar \textit{reward} feedback returned from the environment, which is used to update the Q-table. In this research, the state space, $\boldsymbol{S}$, is composed of the possible forecasting models at the current time:
\begin{equation}
	\boldsymbol{S} = \{\boldsymbol{s}\}=\{s_1, s_2, ..., s_{I}\}
\end{equation}
where $s_i$ means the current forecasting model is M$_i$. $I$ is the number of candidate models. Similarly, the action space, $\boldsymbol{A}$, is composed of the potential forecasting models selected for the next time step:
\begin{equation}
	\boldsymbol{A} = \{\boldsymbol{a}\}=\{a_1, a_2, ..., a_{I}\}
\end{equation}
where $a_j$ means taking the action of switching from the current forecasting model to M$_j$ at the next forecasting time step.

\begin{figure}[!ht]
	\centering
	\includegraphics[width =3.5 in]{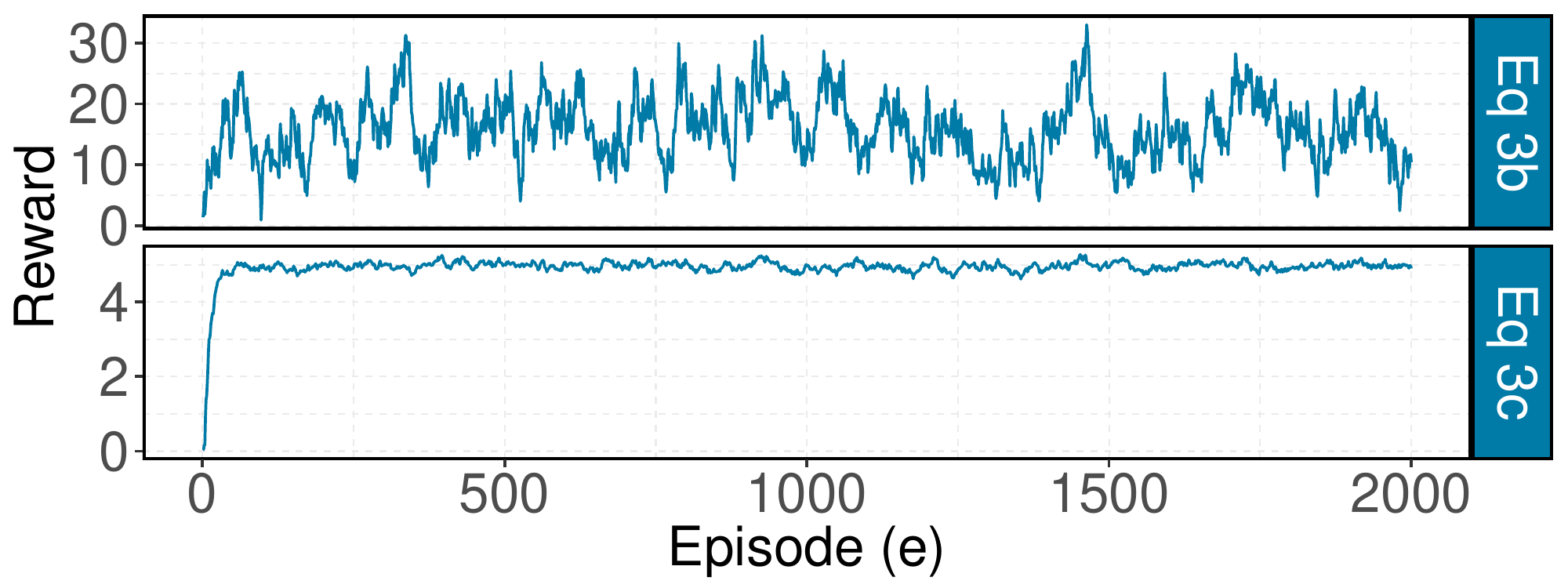}
	\vspace{-.25 in}	
	\caption{Learning curves of two Q-learning training processes with different reward functions. The reward in the vertical axis is defined as the summation of Q-table.}
	\label{learncurve}
\end{figure}

To successfully solve a MDP using Q-learning, the most important step is to maintain a reward matrix, $\boldsymbol{R}$, by a proper reward function, $R(s,a)$. Three reward strategies are considered in this paper, which are based on forecasting error of the next-state model, forecasting error reduction of the next-state model over the current-state model, and the performance ranking improvement of the next-state model over the current-state model (the ranking of the best model is 1). The corresponding reward functions of the three strategies are:
\begin{subequations}
	\begin{align}
	&\label{3a} R^t(s_i, a_j) = \frac{|\hat{y}_j^{(t+1)}-y^{(t+1)}|}{y^{(t+1)}} \\
	&\label{3b} R^t(s_i, a_j) = \frac{|\hat{y}_j^{(t+1)}-y^{(t+1)}|}{y^{(t+1)}} - \frac{|\hat{y}_i^{t}-y^{t}|}{y^{t}}\\
	&\label{3c} R^t(s_i, a_j) = ranking(M_i)-ranking(M_j)
	\end{align}
\end{subequations}
where $\hat{y}_j^t$ is the forecast generated by M$_j$ at time $t$ and $y^t$ is actual value at time $t$. It is found from Eq.~\ref{3a} that the reward function is not related to the current-state if it is only based on the forecasting error of the next-state model (not a MDP), therefore it should be excluded. The second strategy in Eq.~\ref{3b} is able to take the current state into account in the reward function, but the Q-learning algorithm is hard to converge, as shown in the upper part of Fig.~\ref{learncurve}. This is because that the magnitude of forecasting errors do not only depends on forecasting models but is also changing with time. Taking the action of switching from a worse model to the best model might still receive a negative reward (due to forecasting error reduction). Therefore, in this paper, we design the reward function as the model performance ranking improvement, which ensures the effective and efficient convergence of Q-learning, as shown in the lower part of Fig.~\ref{learncurve}.

With state, action, and reward defined, the DMS is realized by training Q-learning agents using the Q-learning training datasets $\boldsymbol{T}_{R\times N}$, which are applied to the DMS process datasets $\boldsymbol{D}_{P\times N}$. The critical component of determining (steps 1-8 in Algorithm~\ref{qlearningalgorithm}) and applying (steps 9-11 in Algorithm~\ref{qlearningalgorithm}) the DMS policy is the Q-table, $\boldsymbol{Q}$, which contains triplets of $s$, $a$, and $Q(s, a)$. As shown in Algorithm~\ref{qlearningalgorithm}, Q values are initialized to be zero and updated repeatedly by Eq.~\ref{qvalue} based on the action reward in the current state and the maximum reward in the next state, where $\alpha$ is the learning rate that controls the aggressiveness of learning, $\gamma$ is a discount factor that weights the future reward. The balance of exploitation and exploration in Q-learning is maintained by adopting a decaying $\epsilon_t$-greedy method~\cite{mnih2013playing}. The Q-learning agent with the decaying $\epsilon_t$-greedy method takes completely random actions  at the beginning, while reduces the randomness with a decaying $\epsilon$ during the learning process. The Q-learning algorithm will eventually converge to the optimal policy, $\boldsymbol{Q}^*$, which is applied to find the optimal actions, $a^*$, in the DMS process.

\begin{algorithm}[ht]
	\caption{Q-learning based Dynamic Model Selection (DMS)}\label{qlearningalgorithm}
	\begin{algorithmic}[1]
		\Require
		\Statex Number of steps, $P$, in a DMS procedure
		\Statex Model pool dimension $N$, number of models pre-selected for DMS $I$
		\Statex Q-learning training dataset $\boldsymbol{T}_{R\times N}$
		\Statex DMS process dataset $\boldsymbol{D}_{P\times N}$
		\Statex Learning rate $\alpha$, discount factor $\gamma$, number of episodes $E$
		\Ensure Select the best model from $N$ models at each step in $\boldsymbol{D}$
		
		\State Initialize $\boldsymbol{Q}=\boldsymbol{\overrightarrow{0}}_{I\times I}$, $\epsilon=1$
		\State Choose the best $I$ models based on $\boldsymbol{T}$
		\For{\texttt{e = 1} \textbf{to} \texttt{E} }
		\State With the probability of $\epsilon$ select a random action $a_e$, otherwise select $a_e=\arg\max\limits_{a\in\boldsymbol{A}}Q^{e}(s^e,a)$
		\State Calculate $\boldsymbol{R}$ by Eq.~\ref{3c}
		\State Update $\boldsymbol{Q}$ by
		\begin{linenomath*}
			\begin{align}\label{qvalue}
			\begin{split}
			&Q^{(e+1)}(s^e,a^e)=(1-\alpha)Q^{e}(s^e,a^e)+\\&\alpha[R^e(s^e,a^e)+\gamma\max\limits_{a\in\boldsymbol{A}}Q^{e}(s^{(e+1)},a)]
			\end{split}
			\end{align}
		\end{linenomath*}
	 \State $\epsilon \leftarrow \epsilon - \frac{1}{n}$
	\EndFor
		\For{\texttt{p = 1} \textbf{to} \texttt{P}}
		\State Take action $a^*_p=\arg\max\limits_{a\in\boldsymbol{A}}Q^*(s^{p},a)$
		\EndFor
	\end{algorithmic}
\end{algorithm}

\subsection{The STLF with Q-learning based DMS}
As shown in Fig.~\ref{framework}, the training data is used for building $N$ machine learning based forecasting models, and the validation data is used to tune forecasting model hyperparameters and Q-learning parameters ($\alpha$, $\gamma$, $E$, $R$, $P$, and $I$). The effectiveness of the developed STLF with Q-learning based DMS framework (M$_Q$) is verified by the testing data. At the forecasting stage (testing stage), a moving window is adopted to update the Q-learning agent and pre-select the best $I$ models based on the recent historical data (Q-learning training dataset $\boldsymbol{T}$ in Algorithm~\ref{qlearningalgorithm}). The Q-learning agent is then used to make DMS for the next $P$ steps. The moving window moves $P$ steps forward and repeats the procedure in Algorithm~\ref{qlearningalgorithm} until the previous DMS process is finished.

\section{Case Studies}\label{section3}
\subsection{Data Description and Q-learning Parameter Setting}
\begin{figure}[!ht]
	\centering
	\includegraphics[width =3.5 in]{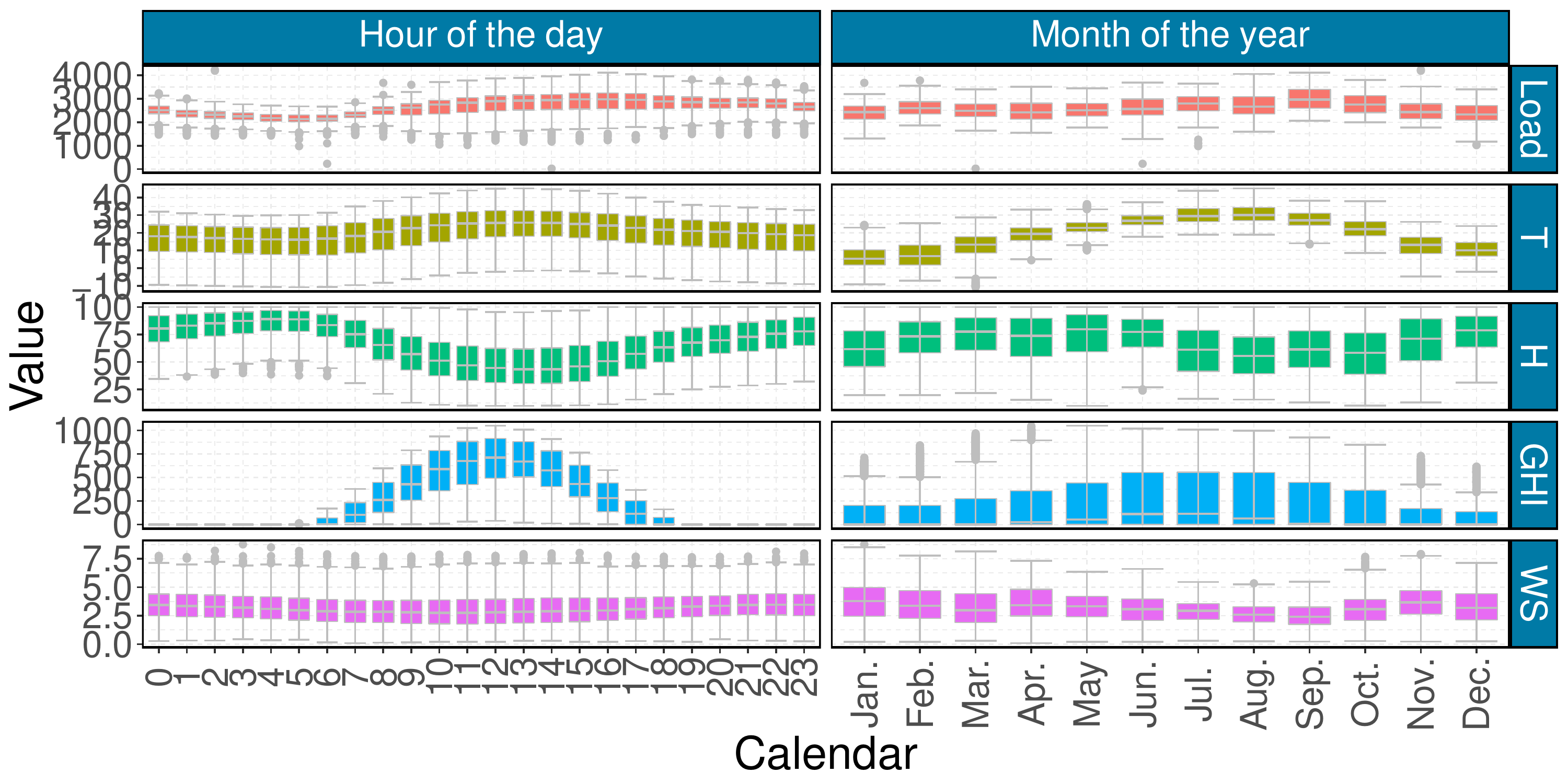}
	\vspace{-.25 in}	
	\caption{Temporal statistics of UTD load and four key weather variables. Lines in the boxes are the medians. The interquartile range box represents the middle 50\% of the data. The upper and lower bounds are maximum and minimum values of the data, respectively, excluding outliers.}
	\label{databoxplot}
\end{figure}

In this paper, hourly campus load data of The University of Texas at Dallas (UTD) and hourly weather data retrieved from the National Solar Radiation Database\footnote{https://nsrdb.nrel.gov} is used for 1-hour-ahead load forecasting. The UTD load includes consumptions of 13 buildings with diverse patterns. The weather data includes meteorological variables such as temperature (T), humidity (H), global horizontal irradiance (GHI), and wind speed (WS). Temporal statistics of the load and four key weather variables are shown in Fig.~\ref{databoxplot}. It is found that the load and weather data is impacted by calendar effects~\cite{feng2018unsupervised, feng2018characterizing}. For example, the load is higher and more chaotic from 8 am to 5 pm, which is possibly due to the working hours of the university. The minimum load in June and December is smaller than that in other months due to the holidays. Therefore, calendar units are also included in the dataset, which are hour of the day, day of the week, and month of the year. 

Both load and weather data spans from January 1\textsuperscript{st} 2014 to December 31\textsuperscript{st} 2015. The data ranging from January 1\textsuperscript{st} 2014 to October 31\textsuperscript{st} 2014 is used to train models in the forecasting model pool, while the data from November 1\textsuperscript{st} 2014 to December 31 \textsuperscript{st} 2014 is used to tune forecasting model hyperparameters. The Q-learning parameters are also determined based on the validation data. Specifically, $\alpha=0.1$ and $\gamma=0.8$, which ensures the learning speed and also respects the future reward. The moving window parameters in Q-learning are set as: $I=4$, $P=4$, $R=72$, which require fewer episodes ($E=100$) to ensure convergence. The data in 2015 is used for testing. 

\subsection{Q-learning DMS Effectiveness}
\begin{figure}[!ht]
	\centering
	\includegraphics[width =3.5 in]{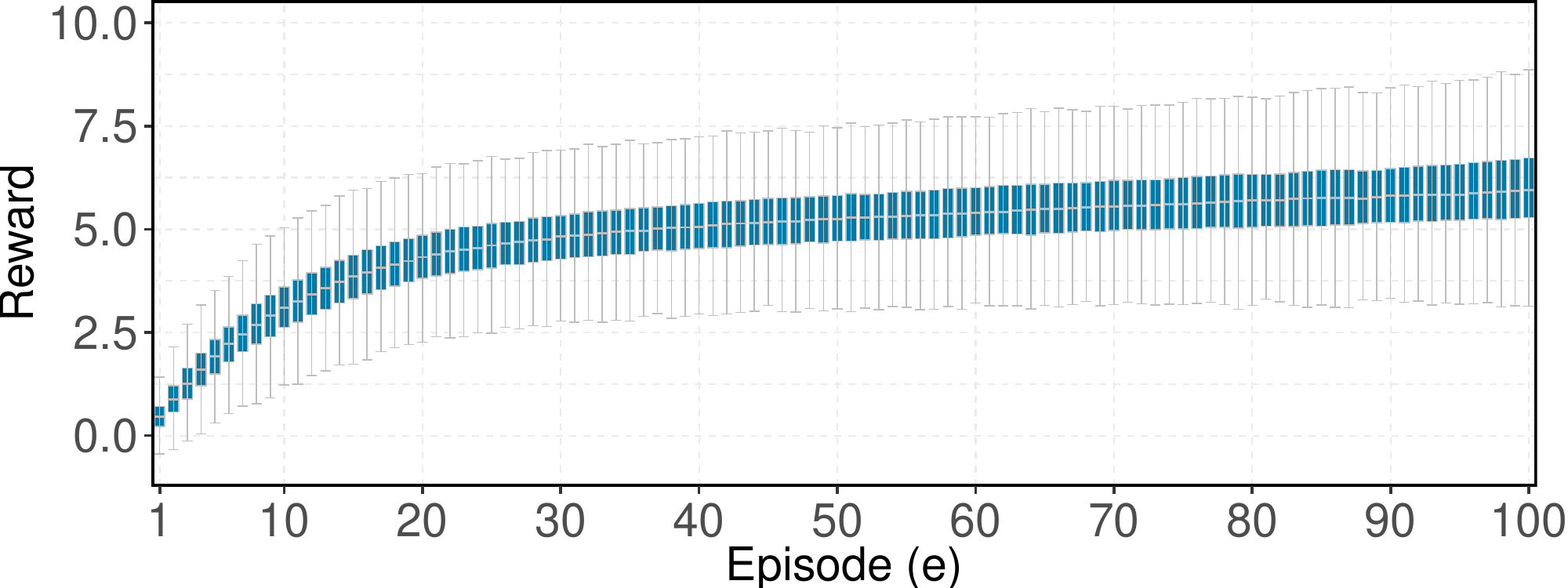}
	\vspace{-.25 in}
	\caption{Learning curve statistics of 2,190 Q-learning agents. The meanings of the boxplot features are same to those in Fig.~\ref{databoxplot}.}
	\label{qconverge}
\end{figure}
The effectiveness of the Q-learning DMS is evaluated based on the testing dataset with 365 days in 2015. Based on the moving window parameters, a Q-learning agent is trained every four time steps to make DMS. Therefore, there are totally 2,190 Q-learning agents built to select proper forecasting models for the 8,760 time steps in 2015. Figure~\ref{qconverge} shows the statistics of Q-learning agent learning curves, which indicates the fast and successful convergence of Q-learning agents. Specifically, Q-learning agents learn extremely fast from interactions with the environment  in the first 30 episodes. After the first 30 episodes, even though the exploration probability is still high ($\epsilon=0.7$ when $e=30$), Q-learning agents learn slowly and tend to converge. Thus, Q-learning agents converge effectively and efficiently in the selected case study.

\begin{figure}[!ht]
	\centering
	\includegraphics[width =3.5 in]{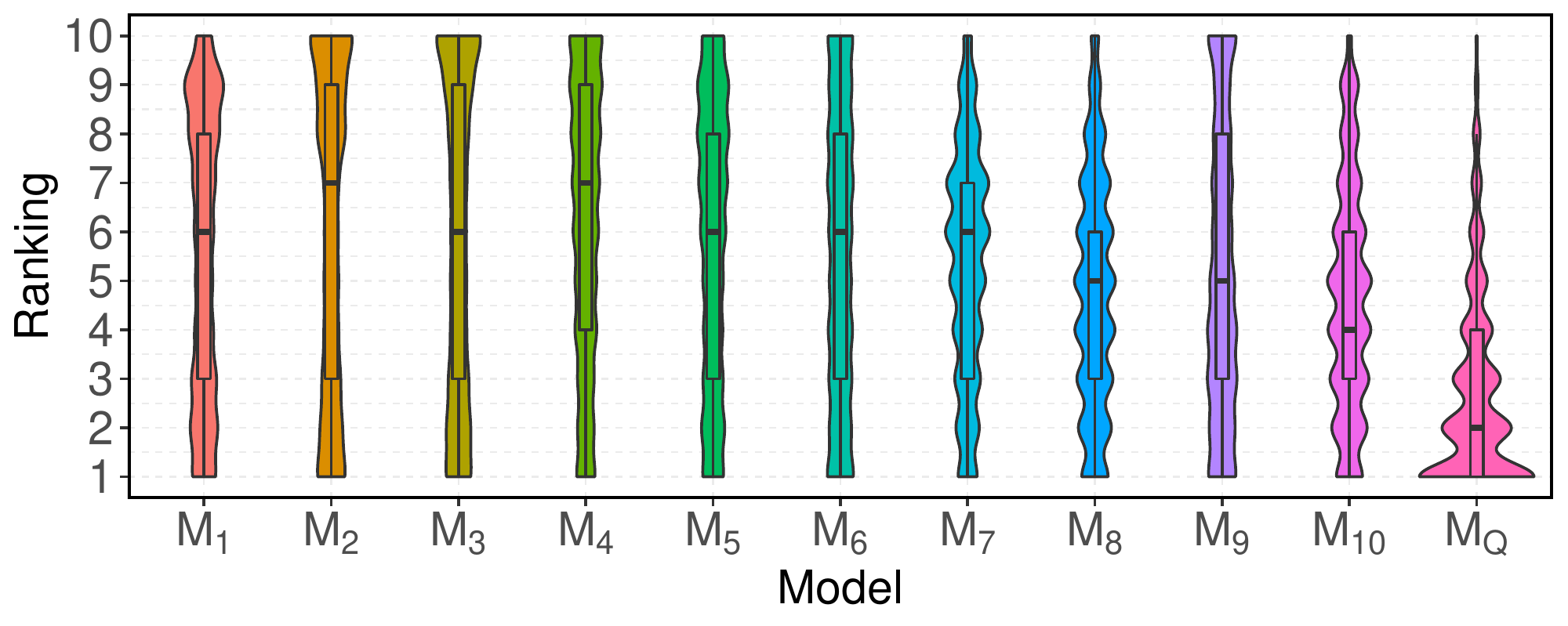}
	\vspace{-.25 in}
	\caption{Violin plot of forecasting model rankings. The changing width of a violin indicates the distribution of rankings of a forecasting model, while the boxplot inside a violin shows similar information as Fig.~\ref{databoxplot}.}
	\label{violinplot}
\end{figure}

To verify the effectiveness of the Q-learning DMS, the rankings of each model at every time step of one year are counted and statistically shown as a violin plot in Fig.~\ref{violinplot}. It is observed from the figure that forecasting models perform distinctively at different time steps, where every model could become the best or the worst at a certain time step. Each model also shows unique characteristics. For example, the three ANN models (M$_1$, M$_2$, M$_3$) rank 8\textsuperscript{th}, 9\textsuperscript{th}, 10\textsuperscript{th} (the worst three) and 1\textsuperscript{st}, 2\textsuperscript{nd} (the best two) more times than other rankings. An SVR model (M$_6$) almost has the same chance for each ranking. It's important to note that no single model (M$_1$-M$_{10}$) dominates others in the STLF. The effectiveness of the Q-learning DMS is evident by comparing the violin of M$_Q$ with violins of other models. It is found that Q-learning agents select a best four models with a 75\% chance and they tend to select a better model, since M$_Q$ model has an upward violin.
\begin{figure}[!ht]
	\centering
	\includegraphics[width =3.5 in]{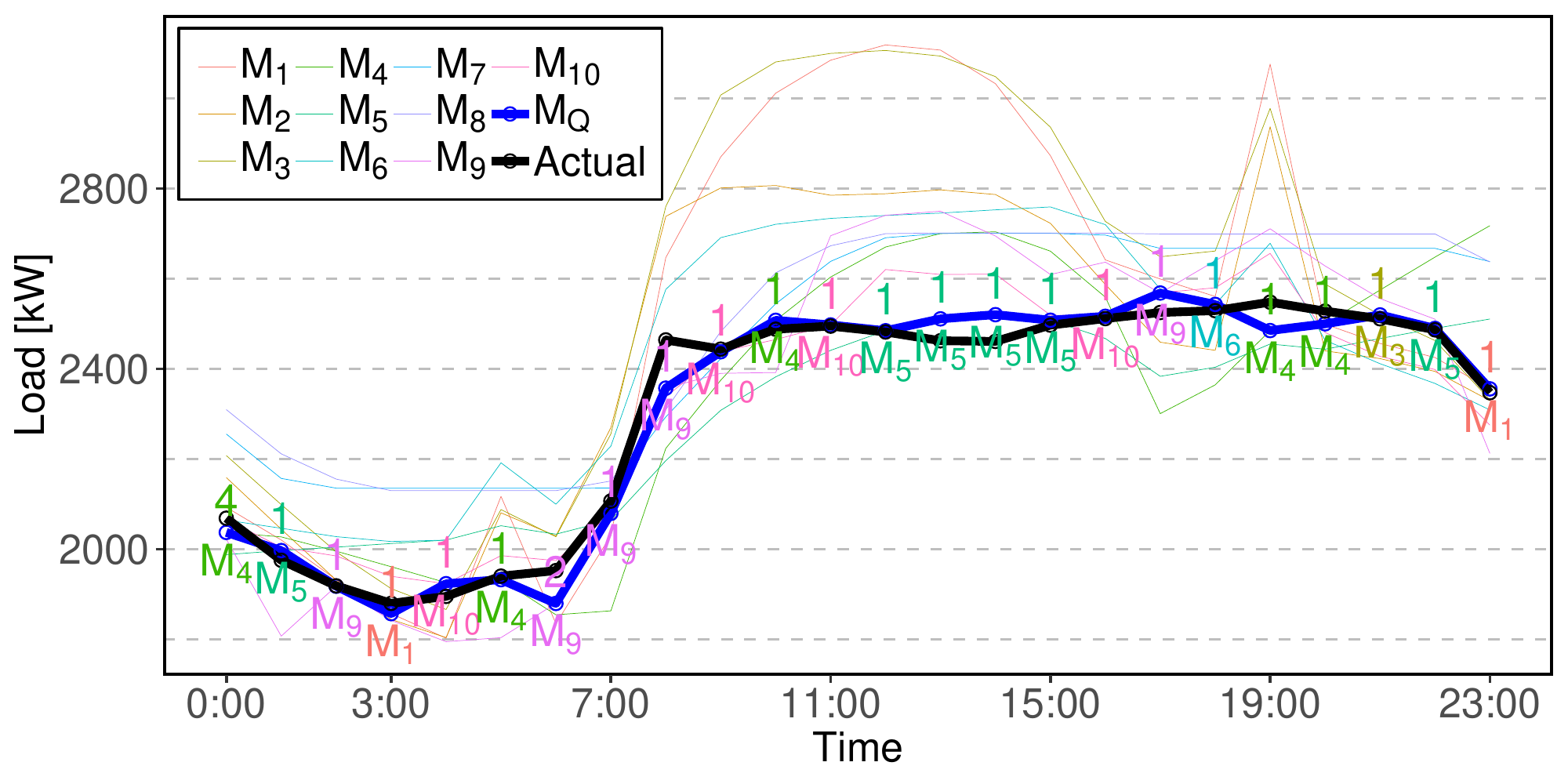}
	\vspace{-.25 in}
	\caption{Forecasting and actual time series of one day; values above forecasting points are rankings of the selected models; symbols below forecasting points are names of the selected models; the annotation font color and the line color of the same model are identical.}
	\label{forets}
\end{figure}

Figure~\ref{forets} shows the actual and forecasting time series of one day. Specifically, the thin lines represent 10 forecasting models in the model pool, the thick blue line represents the M$_Q$, and the thick black line represents the actual load time series. In general, most selected models rank 1\textsuperscript{st} except for the models at 0 am and 7 am. However, the Q-learning agent intends to select the 4\textsuperscript{th} model at 0 am since it will receive a large reward ($R(s_4, a_5)$ = 3) by switching from M$_4$ (ranking 4\textsuperscript{th}) at 0 am to M$_5$ (ranking 1\textsuperscript{st}) at 1 am. It is also logic for the Q-learning agent to select M$_9$ at 7 am because it values the future reward of this selection.

\subsection{Overall Forecasting Accuracy}
To evaluate the forecasting accuracy, two error metrics are employed, which are normalized mean absolute error ($nMAE$) and mean absolute percentage error ($MAPE$)~\cite{feng1short}. Two more metrics are used to compare the developed M$_Q$ model with 10 machine learning models listed in Table~\ref{kernel}, which are $nMAE$ reduction ($Imp^A$) and $MAPE$ reduction ($Imp^P$)~\cite{feng2018unsupervised}. The overall forecasting performance of the developed M$_Q$ model and the 10 forecasting models is summarized in Table~\ref{errortable}. The overall forecasting $nMAE$ and $MAPE$ of the developed M$_Q$ model are 3.23\% and 5.61\%, respectively, which indicates that the average forecasting error is extremely small with respect to both the load capacity and actual load. By comparing the developed M$_Q$ model to the candidate models, it is found that the improvements are significant and consistent. The average $Imp^A$ and $Imp^P$ are 49.50\% and 47.45\%, respectively. Therefore, we can conclude that the developed STLF with Q-learning DMS is effective and outperforms the candidate models.
\begin{table}[!ht]
	\caption{Forecasting $nMAE$ [\%], $MAPE$ [\%], and improvements [\%] of the developed model over other models}
	\label{errortable}
	\begin{center}  
		\begin{tabular}{ccccccccccc}  
			\rowcolor{cerulean}&\multicolumn{2}{c}{\textcolor{white}{\textbf{Forecasting Error}}}&\multicolumn{2}{c}{\textcolor{white}{\textbf{Forecasting Improvement}}}\\
			\rowcolor{cerulean}\multirow{-2}{*}{\textbf{\textcolor{white}{\textbf{Model }}}}&\textcolor{white}{$nMAE$}&\textcolor{white}{$MAPE$}&\textcolor{white}{$Imp^A$}&\textcolor{white}{$Imp^P$}\\
			\multicolumn{1}{l|}{M$_1$}&6.74&11.53&52.06&51.33\\
			\multicolumn{1}{l|}{M$_2$}&7.41&12.71&56.40&55.88\\
			\multicolumn{1}{l|}{M$_3$}&7.81&13.50&\textbf{58.65}&\textbf{58.44}\\
			\multicolumn{1}{l|}{M$_4$}&7.20&11.58&55.16&51.55\\
			\multicolumn{1}{l|}{M$_5$}&6.58&10.77&50.89&47.92\\
			\multicolumn{1}{l|}{M$_6$}&6.35&10.22&49.16&45.11\\
			\multicolumn{1}{l|}{M$_7$}&5.89&9.97&45.14&43.74\\
			\multicolumn{1}{l|}{M$_8$}&5.47&9.37&40.92&40.11\\
			\multicolumn{1}{l|}{M$_9$}&6.20&10.12&47.88&44.54\\
			\multicolumn{1}{l|}{M$_{10}$}&\textbf{5.27}&\textbf{8.74}&38.73&35.84\\
			\multicolumn{1}{l|}{M$_Q$}&\textcolor{grn}{\textbf{3.23}}&\textcolor{grn}{\textbf{5.61}}&NA&NA\\
			\hline
		\end{tabular}
	\end{center}
	\small Note: \textbf{Bold values} indicate the best candidate models or the most improvements of the developed model over candidate models, while \textcolor{grn}{\textbf{bold green values}} indicate the developed M$_Q$ model.
\end{table}

\section{Conclusions and Future Work}\label{section4}
This paper developed a novel short-term load forecasting (STLF) method based on reinforcement learning dynamic model selection (DMS). First, a forecasting model pool that consists of 10 state-of-the-art machine learning based forecasting models was built and generated forecasts with diverse performance. Then, Q-learning agents were trained based on rewards of model ranking improvements. The best forecasting models were selected from candidate models dynamically by the optimal DMS policy. Case studies based on two-year load and weather data showed that:
\begin{itemize}
	\item[(1)] Q-learning agents learned effectively and efficiently from the designed MDP environment of DMS.
	\item[(2)] The developed STLF with Q-learning DMS improved the forecasting accuracy by approximately 50\%, compared to benchmark machine learning models.
\end{itemize}

Future work will focus on applying the Q-learning in predictive distribution selection in probabilistic forecasting and exploring deep reinforcement learning in model or predictive distribution selection.

\bibliographystyle{IEEEtran}
\bibliography{IEEEfull,reinforcementlearning}

\begin{thebibliography}{10}
\providecommand{\url}[1]{#1}
\csname url@samestyle\endcsname
\providecommand{\newblock}{\relax}
\providecommand{\bibinfo}[2]{#2}
\providecommand{\BIBentrySTDinterwordspacing}{\spaceskip=0pt\relax}
\providecommand{\BIBentryALTinterwordstretchfactor}{4}
\providecommand{\BIBentryALTinterwordspacing}{\spaceskip=\fontdimen2\font plus
\BIBentryALTinterwordstretchfactor\fontdimen3\font minus
  \fontdimen4\font\relax}
\providecommand{\BIBforeignlanguage}[2]{{%
\expandafter\ifx\csname l@#1\endcsname\relax
\typeout{** WARNING: IEEEtran.bst: No hyphenation pattern has been}%
\typeout{** loaded for the language `#1'. Using the pattern for}%
\typeout{** the default language instead.}%
\else
\language=\csname l@#1\endcsname
\fi
#2}}
\providecommand{\BIBdecl}{\relax}
\BIBdecl

\bibitem{chen2017short}
Y.~Chen, P.~Xu, Y.~Chu, W.~Li, Y.~Wu, L.~Ni, Y.~Bao, and K.~Wang, ``Short-term
  electrical load forecasting using the support vector regression ({SVR}) model
  to calculate the demand response baseline for office buildings,''
  \emph{Applied Energy}, vol. 195, pp. 659--670, 2017.

\bibitem{saksornchai2005improve}
T.~Saksornchai, W.-J. Lee, K.~Methaprayoon, J.~R. Liao, and R.~J. Ross,
  ``Improve the unit commitment scheduling by using the neural-network-based
  short-term load forecasting,'' \emph{IEEE Transactions on Industry
  Applications}, vol.~41, no.~1, pp. 169--179, 2005.

\bibitem{cui2017characterizing}
M.~Cui, J.~Zhang, C.~Feng, A.~R. Florita, Y.~Sun, and B.-M. Hodge,
  ``Characterizing and analyzing ramping events in wind power, solar power,
  load, and netload,'' \emph{Renewable Energy}, vol. 111, pp. 227--244, 2017.

\bibitem{shi2017deep}
H.~Shi, M.~Xu, and R.~Li, ``Deep learning for household load forecasting--a
  novel pooling deep {RNN},'' \emph{IEEE Transactions on Smart Grid}, 2017.

\bibitem{raza2015review}
M.~Q. Raza and A.~Khosravi, ``A review on artificial intelligence based load
  demand forecasting techniques for smart grid and buildings,'' \emph{Renewable
  and Sustainable Energy Reviews}, vol.~50, pp. 1352--1372, 2015.

\bibitem{yildiz2017review}
B.~Yildiz, J.~Bilbao, and A.~Sproul, ``A review and analysis of regression and
  machine learning models on commercial building electricity load
  forecasting,'' \emph{Renewable and Sustainable Energy Reviews}, vol.~73, pp.
  1104--1122, 2017.

\bibitem{alamaniotis2012evolutionary}
M.~Alamaniotis, A.~Ikonomopoulos, and L.~H. Tsoukalas, ``Evolutionary
  multiobjective optimization of kernel-based very-short-term load
  forecasting,'' \emph{IEEE Transactions on Power Systems}, vol.~27, no.~3, pp.
  1477--1484, 2012.

\bibitem{feng2018short}
C.~Feng and J.~Zhang, ``Short-term load forecasting with different aggregration
  strategies,'' in \emph{ASME 2018 International Design Engineering Technical
  Conferences and Computers and Information in Engineering Conference}.\hskip
  1em plus 0.5em minus 0.4em\relax American Society of Mechanical Engineers,
  2018.

\bibitem{wang2016factors}
X.~Wang, W.-J. Lee, H.~Huang, R.~L. Szabados, D.~Y. Wang, and P.~Van~Olinda,
  ``Factors that impact the accuracy of clustering-based load forecasting,''
  \emph{IEEE Transactions on Industry Applications}, vol.~52, no.~5, pp.
  3625--3630, 2016.

\bibitem{feng2017data}
C.~Feng, M.~Cui, B.-M. Hodge, and J.~Zhang, ``A data-driven multi-model
  methodology with deep feature selection for short-term wind forecasting,''
  \emph{Applied Energy}, vol. 190, pp. 1245--1257, 2017.

\bibitem{yan2017q}
J.~Yan, H.~He, X.~Zhong, and Y.~Tang, ``Q-learning-based vulnerability analysis
  of smart grid against sequential topology attacks,'' \emph{IEEE Transactions
  on Information Forensics and Security}, vol.~12, no.~1, pp. 200--210, 2017.

\bibitem{mnih2013playing}
V.~Mnih, K.~Kavukcuoglu, D.~Silver, A.~Graves, I.~Antonoglou, D.~Wierstra, and
  M.~Riedmiller, ``Playing atari with deep reinforcement learning,''
  \emph{arXiv preprint arXiv:1312.5602}, 2013.

\bibitem{feng2018unsupervised}
C.~Feng, M.~Cui, B.-M. Hodge, S.~Lu, H.~F. Hamann, and J.~Zhang, ``An
  unsupervised clustering-based short-term solar forecasting methodology using
  multi-model machine learning blending,'' \emph{arXiv preprint
  arXiv:1805.04193}, 2018.

\bibitem{feng2018characterizing}
C.~Feng, M.~Sun, M.~Cui, E.~K. Chartan, B.-M. Hodge, and J.~Zhang,
  ``Characterizing forecastability of wind sites in the united states,''
  \emph{Renewable Energy}, 2018.

\bibitem{feng1short}
C.~Feng, M.~Cui, M.~Lee, J.~Zhang, B.-M. Hodge, S.~Lu, and H.~F. Hamann,
  ``Short-term global horizontal irradiance forecasting based on sky imaging
  and pattern recognition,'' in \emph{IEEE PES General Meeting}.\hskip 1em plus
  0.5em minus 0.4em\relax IEEE, 2017.

\end{thebibliography}
\end{document}